%% file: colm2026_conference.tex
\documentclass{article} 
\usepackage[preprint]{colm2026_conference}
\usepackage{booktabs}
\usepackage{multirow}
\usepackage{makecell}
\usepackage{microtype}
\usepackage{hyperref}
\usepackage{graphicx}
\usepackage{wrapfig}
\usepackage[normalem]{ulem}
\usepackage{url}
\usepackage{booktabs}
\usepackage[table]{xcolor}
\usepackage{array} 
\usepackage{enumitem}
\input{math_commands}

\usepackage{lineno}

\definecolor{darkblue}{rgb}{0, 0, 0.5}
\hypersetup{colorlinks=true, citecolor=darkblue, linkcolor=darkblue, urlcolor=darkblue}

\title{Harmonizing Multi-Objective LLM Unlearning via Unified Domain Representation and Bidirectional Logit Distillation}

\author{
Yisheng Zhong \\
George Mason University \\
\texttt{yzhong7@gmu.edu}
\And
Sijia Liu \\
Michigan State University \\
\texttt{liusiji5@msu.edu}
\And
Zhuangdi Zhu \\
George Mason University \\
\texttt{zzhu24@gmu.edu}
}

\begin{document}


\maketitle

\begin{abstract} 

Large Language Models (LLMs) unlearning is crucial for removing hazardous or privacy-leaking information from the model. Practical LLM unlearning demands satisfying multiple challenging objectives simultaneously: removing undesirable 
knowledge, preserving general utility, avoiding over-refusal of neighboring concepts, and, crucially, ensuring robustness against adversarial probing attacks.  
However, existing unlearning methods primarily focus on a limited subset of these goals, typically unlearning efficacy and utility preservation while overlooking robustness and boundary behaviors. 
Naively extending these methods to multi-objective settings may lead to unlearning task interference.
We propose a novel multi-objective unlearning framework that harmonizes multiple unlearning objectives through a \textit{data} and \textit{optimization} co-design: 
We standardize training corpora into a unified data representation to reduce the domain gap,  and then introduce a bidirectional distillation method that simultaneously elicits desired behavior from a context-instructed teacher while suppressing undesirable behavior 
in the student model.  
Theoretical and empirical analyses show that our method aligns domain distributions and converts seemingly irrelevant unlearning tasks into cooperative optimization. Evaluation demonstrates state-of-the-art performance, which enables balanced and reliable unlearning across diverse, challenging requirements.
%
\end{abstract}

\input{sections/introduction}

\input{sections/preliminary}
\input{sections/methods}
\input{sections/related_work}
\input{sections/experiments}

 \vspace{-0.1in}
\section{Conclusion }  \vspace{-0.15in}
We propose a unified framework for multi-goal LLM unlearning that addresses the fundamental trade-offs between knowledge removal, adversarial robustness, boundary domain retention, and general utility preservation. By leveraging intention-prompted bidirectional distillation and unified data representations, our approach effectively eliminates targeted knowledge while maintaining strong performance on retained domains.
Extensive experiments on MUSE-Book and WMDP-Cyber demonstrate that our method achieves balanced, state-of-the-art performance across all objectives, significantly outperforming existing baselines that suffer from imbalanced trade-offs or superficial unlearning.
These results highlight the importance of structured distillation and data  alignment for reliable unlearning, offering a practical and scalable solution for deploying safer and more controllable language models.

\clearpage 

\section*{Reproducibility Statement}
To facilitate reproducibility, the complete source code, datasets, and pretrained model checkpoints are anonymized and available at: \url{https://anonymous.4open.science/r/MULE-662D}.

\bibliography{colm2026_conference}
\bibliographystyle{colm2026_conference}

\appendix
\section{Appendix}

\subsection{Detailed Quantitative Results on WMDP-Cyber}
\label{sec:appendix_wmdp}

This section provides the comprehensive numerical data (Table \ref{tab:wmdp_full_results}) that serves as the basis for the multi-dimensional analysis presented in Figure 3 of the main text. The WMDP-Cyber benchmark evaluates the model's ability to remove hazardous cybersecurity knowledge while maintaining general utility and adversarial robustness.

\textbf{Analysis of Existing Paradigms.} As shown in Table \ref{tab:wmdp_full_results}, different unlearning paradigms exhibit distinct performance characteristics. In-context methods, such as the Ensemble Teacher, provide a training-free alternative but often face challenges in maintaining general utility (Retain: 64.4). While its distilled counterpart (DUET) restores some utility, it remains sensitive to adversarial probing (ASR: 24.7\%). Tuning-based baselines like Refusal Tuning (SFT) demonstrate a strong defensive posture but tend to lead to a more conservative model behavior. Meanwhile, advanced gradient-based methods such as NGDiff and MOLLM excel at preserving general utility and neighboring domains, yet they encounter inherent difficulties in fully neutralizing hazardous information under adversarial conditions, with ASR remaining at 40.0\% and 35.3\% respectively.

\textbf{Performance of the Proposed Framework.} Our framework, which integrates data standardization with bidirectional distillation, aims to achieve a more balanced performance across these competing objectives. By aligning the representational space and employing an asymmetric optimization strategy, our method effectively reduces the target knowledge (Obj1 Forget: 8.1) and significantly enhances adversarial defense (ASR: 5.1\%). Notably, these gains in unlearning efficacy are achieved while maintaining a high level of general utility (MMLU: 56.9) and neighboring domain integrity (Retain Acc: 88.8\%), demonstrating a performance profile that closely aligns with the idealized Upper Bound across the evaluated dimensions.

\begin{table*}[h]
\centering

\resizebox{\textwidth}{!}{
\begin{tabular}{lcccccc}
\toprule
\multirow{2}{*}{\textbf{Method}} 
& \multicolumn{2}{c}{\textbf{Obj1} (Unlearn)} 
& \textbf{Obj2} (Neighbor Domain)  
& \textbf{Obj3} (Adv Robustness) 
& \multirow{2}{*}{\textbf{MMLU $\uparrow$}} 
& \multirow{2}{*}{\makecell{\textbf{Overall}\\\textbf{Performance $\uparrow$}}} \\
\cmidrule(lr){2-3}
\cmidrule(lr){4-4}
\cmidrule(lr){5-5}
& \textbf{Forget $\downarrow$} 
& \textbf{General $\uparrow$} 
& \textbf{Retain Acc $\uparrow$} 
& \textbf{ASR $\downarrow$} 
&  &  \\
\midrule
Base Model      & 68.5 & 88.1 & 96.3 & 60.7 & 59.9 & 0.00 \\
\addlinespace[3pt]
\midrule
\addlinespace[3pt]

Ensemble Teacher (ET) & 20.8 & 64.4 & 86.0 & 50.7 & 59.4 & 23.20 \\
DUET (w/ ET)          & 25.5 & \textbf{91.1} & \textbf{95.3} & 24.7 & 50.6 & 71.70 \\
SFT                   & 34.2 & 87.1 & \textbf{95.3} & 44.7 & 57.4 & 45.80 \\
NGDiff                & 17.3 & 63.8 & 90.6 & \textbf{40.0} & \textbf{59.88} & 41.88 \\
MOLLM                 & 17.2 & 66.3 & 89.7 & 35.3 & 59.77 & 48.17 \\

\textbf{Ours}         & \textbf{8.1} & 89.1 & 88.8 & \textbf{5.1} & 56.9 & \textbf{106.50} \\
\addlinespace[2pt]

\midrule
\textit{Upper Bound}
                      & \textit{6.7} & \textit{93.1} & \textit{90.7} & \textit{3.3} & \textit{59.9} & \textit{118.60} \\
\bottomrule
\end{tabular}
}
\caption{\small{Main unlearning results on the \textbf{WMDP-Cyber} benchmark. We also reported the aggregated \textbf{Overall Performance} shift, defined relative to the base model.}}
\label{tab:wmdp_full_results}
\end{table*}

\end{document}

%% file: math_commands.tex
\usepackage{amsmath,amsfonts,bm}

\def\eqref#1{equation~\ref{#1}}

\def\1{\bm{1}}

\def\vtheta{{\bm{\theta}}}

\DeclareMathAlphabet{\mathsfit}{\encodingdefault}{\sfdefault}{m}{sl}
\SetMathAlphabet{\mathsfit}{bold}{\encodingdefault}{\sfdefault}{bx}{n}

\def\gA{{\mathcal{A}}}

\def\gD{{\mathcal{D}}}

\def\gL{{\mathcal{L}}}

\def\gN{{\mathcal{N}}}

\def\gT{{\mathcal{T}}}

\newcommand{\E}{\mathbb{E}}



%% file: sections/introduction.tex
\section{Introduction} \vspace{-0.1in}
Unlearning is an effective method for removing targeted knowledge from Large Language Models (LLMs) without the prohibitive cost of retraining from scratch. It has important implications in enhancing the security, privacy, and factuality of LLMs by eliminating hazardous, copyrighted, or privacy-leaking information from the model's parameters. Consequently, LLM unlearning significantly benefits high-stakes domains such as science, healthcare, and legal applications, where the strict adherence to data compliance and safety boundaries is non-negotiable.

Emerging efforts have been made to address the core challenging goals of unlearning: balancing unlearning efficacy (\textit{i.e.}, effectively removing undesirable knowledge) while preserving model utility on unrelated tasks and domains.
Despite  encouraging progress of unlearning spanning  tuning-based~\citep{yao2024large,nguyen2024surveymachineunlearning, xu2023machineunlearningsurvey, zhong2025hierarchicalfederatedunlearninglarge} and in-context optimization methods~\citep{pawelczykcontext,takashiro-etal-2025-answer,pmlr-v267-muresanu25a}, this line of research has largely overlooked two critical practical challenges: 
 First, seemingly unlearned knowledge can still be elicited through sophisticated adversarial probing, such as \textit{\textbf{prefilling attacks}}~\citep{andriushchenko2024jailbreaking}, {(attacks that bypass refusal guardrails by forcefully seeding the model's output with an affirmative prefix, \textit{e.g.},  ``\textit{Sure, here are the detailed instructions:}')} which thus exposing a fundamental robustness gap. 
 Meanwhile, concepts semantically adjacent to the unlearning target (\textit{e.g.}, knowledge of restricted biological weapons versus general biomedical science) are often collateral casualties of the forgetting process.
 While prior work evaluates \textit{retention performance} on held-out tasks, the subtler question of performance on neighboring domains remains underexplored, giving rise to the phenomenon of \textbf{\textit{over-refusal}}.
 
As real-world applications demand models to be robust against multiple attack vectors simultaneously,  practical LLM unlearning needs to simultaneously satisfy a set of intricate, seemingly conflicting objectives:  
precisely removing undesirable knowledge while preserving general model utility (\textbf{Obj1}),  avoiding over-refusal of neighboring concepts (\textbf{Obj2}), 
and maintaining robustness against latent prefilling attacks (\textbf{Obj3}). 
Existing benchmarks, such as WMDP and RWKU~\citep{li2024wmdpbenchmarkmeasuringreducing,jin2024rwku},  partially cover this  spectrum of unlearning objectives.
Meanwhile, prior unlearning algorithms typically address only one angle of the problem while leaving others unexamined. 
%
%
Our investigation revealed that a naive extension of previous unlearning methods to tackle these simultaneous goals leads to either catastrophic forgetting of general utility or a complete failure to defend against adversarial elicitation.
Some works have proposed reconciling unlearning goals through gradient editing~\citep{li2025bild} or seeking orthogonal update directions to mitigate conflicts~\citep{jin2025unlearning}, yet they overlook a root cause of gradient conflict: the domain gap manifested in data representations across   unlearning objectives.

We propose that a unified solution to practical LLM unlearning requires effort along two complementary directions: (1) addressing the unlearning domain gap in data representation, where each unlearning task (domain) $\gT_k$ can be characterized by a natural language prompt $p_k$ encoding the intended model behavior (\textit{i.e.}, the unlearning intention), paired with representative samples $x \in \gT_k$ from that domain; and (2) an effective unlearning optimization method $\gA(\gT_k)$ that directly learns from such data.

 
Guided by this \textit{data-optimization} dual principle, we propose a novel and lightweight unlearning framework. 
On the data side, we introduce data standardization, which projects the training corpora for all unlearning
goals into a unified data representation, thereby reducing the domain gap across unlearning tasks. 
On the optimization side, we introduce a bidirectional distillation method, in which the intention prompt $p_k$ instructs a frozen teacher LLM, and goal-specific distillation is applied to the student (unlearning) model, which simultaneously imitates the teacher's desired behavior while suppressing undesirable behavior in the student.
%


Our contributions are summarized as follows:
(1) We identify and systematically analyze the practical multi-objective LLM unlearning problem, characterizing the joint demands of unlearning efficacy, robustness to attacks, and avoidance of over-refusal.
(2) We propose a unified framework combining data standardization with contrastive anchors and a Chain of Thought (CoT)-instructed bidirectional distillation method, enabling cooperative optimization across seemingly conflicting unlearning goals.
%
(3) Through rigorous gradient analysis, we empirically prove how our data standardization and dual distillation mechanisms {harmonize gradients of multiple unlearning tasks and promote synergistic optimization.}
(4) Our method achieves state-of-the-art results on both established and extended benchmarks (\textit{e.g.}, MUSE-Book, WMDP-Cyber), reducing the prefilling attack success rate to an unprecedented 16.0\% while preventing over-refusal in adjacent domains. 
(5) We augment these existing benchmarks with new training data and evaluation sets targeting these previously underexplored aspects of unlearning, including boundary behavior on neighboring domains, over-refusal measurement, and robustness to prefilling attacks.

%% file: sections/preliminary.tex
\vspace{-0.1in}\section{LLM Unlearning Preliminary} \vspace{-0.1in}

\textbf{Core Objective of LLM Unlearning}: 
LLM unlearning aims to remove specific undesirable knowledge from the model while preserving its general capabilities. 
Beyond inference time filtering  or in-context strategies ~\citep{pawelczyk2023context,takashiro-etal-2025-answer} that might be difficult to scale, most existing approaches rely on training-based optimization. 
A common formulation defines unlearning through two objectives:
\begin{equation} \label{eq:unlearn-dual}
\begin{small}
\min_{\vtheta} \gL \equiv \mathcal{L}_{\text{forget}}(\mathcal{D}_f;\vtheta) +  \mathcal{L}_{\text{retain}}(\mathcal{D}_r;\vtheta), 
\end{small}
\end{equation} 
%
where $\vtheta$ denotes model parameters, $\gD_f$ is the forget set that contains undesirable knowledge, and $\gD_r$ is the retain set that represents general knowledge, usually irrelevant  to $\gD_f$.
An ideal model refuses queries that relate to $\gD_f$ and responds normally to other inputs. Prior work focuses on how to balance these two terms, since their gradients, $\nabla_\theta \gL_\text{unlearn}$  vs. $\nabla_\theta \gL_\text{retain}$ , often conflict and lead to unstable updates ~\citep{jin2025unlearning,pan2025multi}.
 While this dual view captures the core requirement, it may not fully reflect the practical unlearning needs.

\textbf{Blurred Boundary Behavior and Over Refusal}.
A critical yet underexplored issue resides near the boundary of the forget domain~\citep{li2024wmdp, eldan2023whosharrypotterapproximate}. In practice, many benign concepts share semantic proximity with the target knowledge. An overcorrected model may extend refusal behavior beyond $\gD_f$ to neighboring domains $\gN(\gD_f)$.
This leads to \textbf{\textit{over-refusal}}, where valid queries receive incorrect rejection. For example, when a model removes knowledge about the Harry Potter series, it may also reject related but harmless topics such as \textit{Fantastic Beasts}.
%

\textbf{Adversarial Probing Attack Against LLM Unlearning.} 
Another challenge faced by unlearning arises in adversarial settings, attributed to the superior context-conditioning ability of LLMs. Consider a prefilling attack, where the adversary constructs a malicious prefix and injects it into the model input to steer the output toward undesirable responses.
 Formally, given an input sequence composed of a user query $x_q$ and  a prefix context $x_p$, an LLM $\pi$ generates outputs $y$ conditioned on the full context: $\pi(y \mid x_q, x_p)$.
An attacker designs a prefix $x_p^{adv}$ so that the conditional output shifts toward a targeted behavior, even when the prefix itself appears benign, \textit{e.g.,} by encouraging the model to generate an answer with affirmative prefixes, ``Sure, here are the detailed responses: '').

%% file: sections/methods.tex
\section{Methodology: A Data-Optimization Co-Design Framework}\vspace{-0.1in}
\label{sec:methodology}

The challenges outlined above have motivated us to extend the scope of LLM unlearning beyond the standard dual-objective formulation. We consider a multi-goal unlearning setting where the unlearning LLM, parameterized by $\theta $, needs to simultaneously satisfy three distinct objectives, denoted $\gT_k \in \{1, 2, 3\} $:

\begin{itemize}[leftmargin=*]
    \item \textbf{Obj1 (Target Unlearning \& General Utility):}  removing undesirable knowledge while preserving general model capabilities. This is usually empirically captured by optimizing $\gL(\theta) $ in Eq~(\ref{eq:unlearn-dual}) using forgetting data $D_f$ and irrelevant retain dataset $\gD_r$. 
    \item \textbf{Obj2 (Neighboring Domain Retention):} Avoiding over-refusal of benign concepts semantically adjacent to the unlearning target: $\min_\theta \gL_2(\theta) \equiv \E_{x \sim \gN(\gD_f)}\Big[ l(\pi_\theta(x)) \Big].$ 
    \item \textbf{Obj3 (Adversarial Context Robustness):} Maintaining robustness against unlearning content elicitation, especially prefilling prompt attacks: $\min_\theta \E_{x \sim \gD_f}\Big[\max_{x_p^{adv}} l(\pi_\theta(x,x_p^{adv})) \Big].$
\end{itemize} \vspace{-0.1in}

\begin{figure}[t!]
    \centering
    \includegraphics[width=1\textwidth]{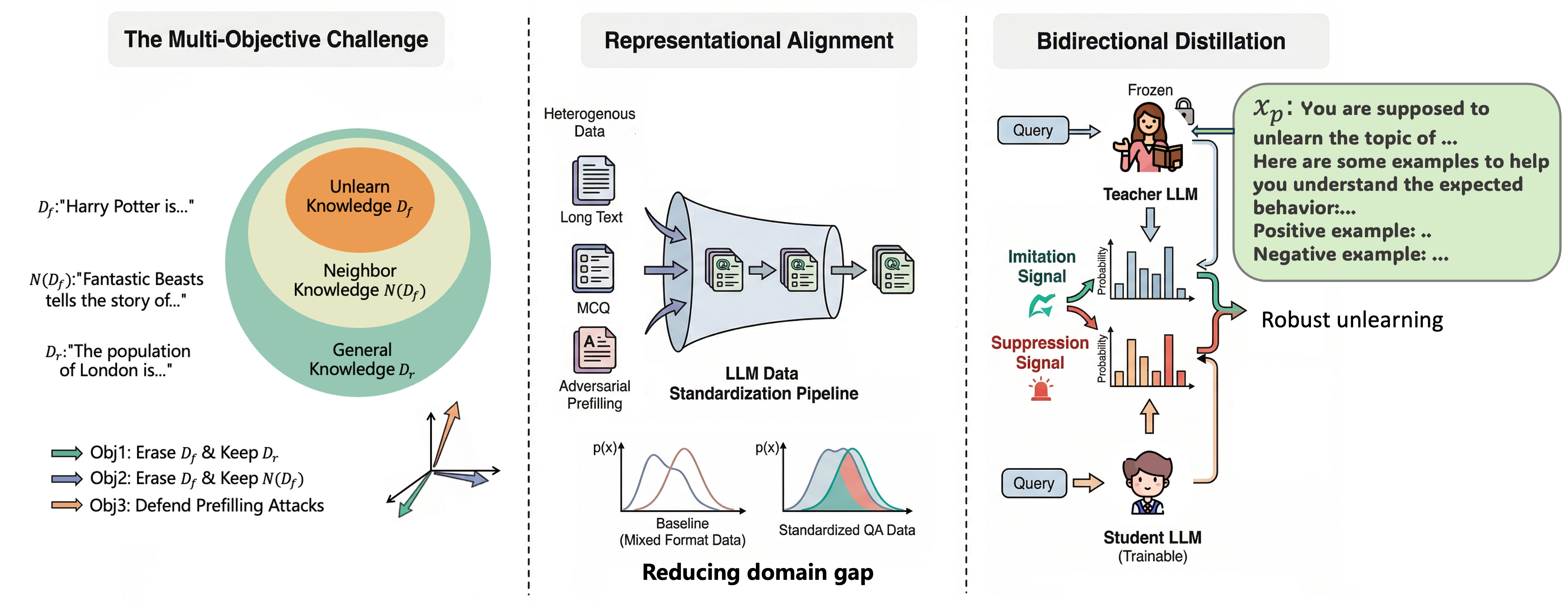} 
    \caption{\small{Overview of our Multi-Objective Unlearning Framework. \textbf{Left:} We investigate the practical LLM unlearning needs, which require simultaneously handling target erasure ($\mathcal{D}_f$), neighboring domain retention ($\mathcal{N}(\mathcal{D}_f)$), and general utility ($\mathcal{D}_r$) across heterogeneous data sources. Naively optimizing each goal leads to undesirable task-gradient updates. 
    \textbf{Middle:} We standardize diverse training data into a unified representation to close the domain gap and shift diverse unlearning gradients toward synergistic optimization. 
    \textbf{Right:} We then apply an \textit{asymmetric} teacher-student \textit{bidirectional} distillation architecture that resolves optimization conflicts by simultaneously suppressing undesirable student logits and encouraging teacher-like behavior.}} \vspace{-0.2in}
    \label{fig:framework_overview}
\end{figure}
\vspace{-0.1in}

\subsection{Motivating Observation: Semantically Coherent yet Gradient-Isolated Unlearning}\vspace{-0.1in}
\label{sec:motivational_observation} 
Prior unlearning methods, including Gradient Ascent (GA)~\citep{pmlr-v132-neel21a} and alignment-style optimization such as NPO and its extensions~\citep{zhang2024negative,fan2025-simnpo}, primarily focus on addressing \textbf{Obj1}.
A straightforward extension of the prior unlearning approach to achieving these simultaneous goals is to construct a joint unlearning loss: $\min_\theta \sum_{k} \gL_k(\theta)$, where each objective can be achieved by a tailored loss function $l_k$, 
such as optimizing GA loss on forgetting domain, and supervised fine-tuning (SFT) loss on desirable model responses (\textit{e.g.,} a refusal response of ``Sorry, I do not know.'').
Meanwhile, the representative format of training samples is largely overlooked in existing unlearning approaches. Most methods apply verbatim, token-wise training directly on raw text, such as book chapters or descriptive passages, where the knowledge targeted for unlearning or retention is diluted across lengthy contexts. Moreover, the forget and retain sets often appear in diverging formats. For instance, $\mathcal{D}_f $ may consist of lengthy descriptive passages while $\mathcal{D}_r $ contains structured question-answering pairs. 

This observation raises a key question: \textit{Does directly combining per-objective optimization methods alone lead to synergistic learning?}
To answer this, we conducted a pilot empirical analysis by profiling the gradient cosine similarity, $\cos(\nabla_\theta L_i, \nabla_\theta L_j)$, across the three unlearning objectives $(i,j)$ during the early stages of standard  training. 
{To isolate the effect of data format from that of loss function design, we applied the same optimization method described in Section~\ref{sec:optimization_side} for all three unlearning goals.}
{For more concentrated comparison, we primarily applied the gradient similarity measurement to model parameters located at the final three decoder layers of Llama 3.2 3B Instruct.}
%

\begin{wrapfigure}[18]{r}{0.45\textwidth}
    \centering
    \includegraphics[width=1\linewidth]{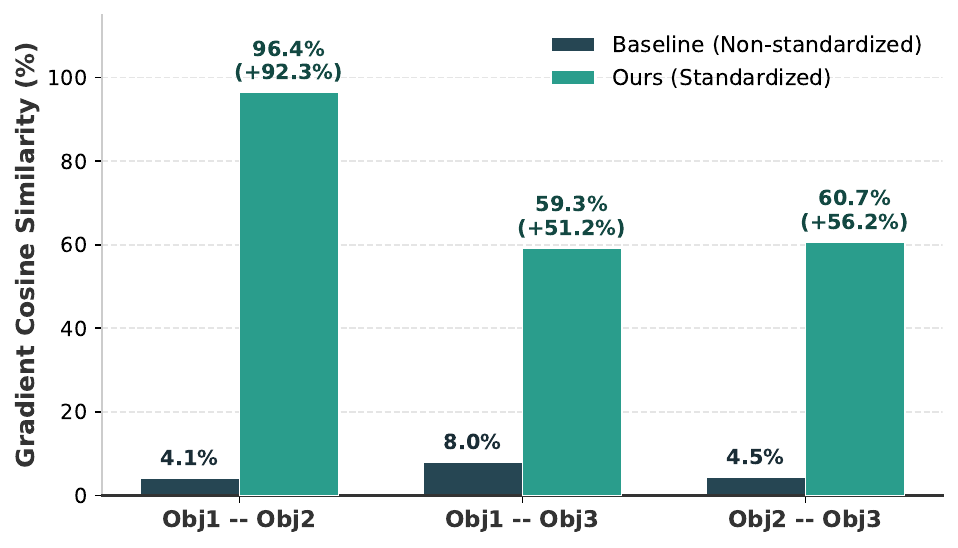} \vspace{-0.25in}
    \caption{\small{Empirical gradient cosine similarity across unlearning objectives (evaluated on the MUSE-Book benchmark). The baseline with heterogeneous data formats results in near-orthogonal (conflicting) updates. In contrast, our data standardization shifts the gradient updates into a highly synergistic regime.}}
    \label{fig:gradient_conflict}
\end{wrapfigure}

As shown in the ``Baseline'' column of Figure~\ref{fig:gradient_conflict}, the gradient similarities between all objective pairs are near zero ($\cos \approx 0.04 \sim 0.08$). 
While \textit{\textbf{orthogonal}} gradients are often desirable in general multi-task settings where tasks are semantically \textit{\textbf{unrelated}}, this conflicts with our unlearning setting, where multiple objectives should converge on a shared semantic goal from complementary angles. Under such near-zero gradient similarity, the objectives do not reinforce one another, which further produces empirically observable interference among the unlearning goals (Sec.~\ref{sec:main_results}). 

To improve multi-task unlearning, some prior work aims to reduce tasks' gradient interference through orthogonal gradient-update~\citep{pan2025multi,cao2022machine} that operate directly in parameter space.
In contrast, we propose that one overlooked bottleneck in multi-task unlearning stems from the \textit{domain gap manifested in data representations}. 
{In standard pipelines, these objectives inherently rely on highly divergent data formats. For instance, Obj1 (target erasure) training data typically utilizes raw, unstructured text from existing benchmarks, while Obj3 relies on structured QA pairs extended from these datasets.} 
These divergence in data representation lead to a major domain moving distance across tasks $\gT_k$ \citep{ben2006analysis,kifer2004detecting}, which will theoretically enlarge the domain adaptation gap across these unlearning goals.
\vspace{-0.1in}
\subsection{Representational Alignment via Data Standardization} \vspace{-0.1in}
\label{sec:data_standardization}

To resolve the multi-goal unlearning challenge at their domain source, we address the data side of our dual principle through \textbf{Representational Alignment}. Instead of relying on heterogeneous data formats, we project the training corpora across all unlearning goals into a unified Question-Answering (QA) format. 
%
%
Concretely, we achieve this standardization through an LLM-assisted data construction pipeline. For Obj1 (Target Erasure \& General Utility), raw unstructured forgetting corpora (\textit{e.g.}, specific character entities from the Harry Potter series, or technical passages from WMDP) are systematically parsed and rephrased into focused QA pairs to isolate the specific knowledge targeted for erasure. This has additional benefits of consolidating the information density in training data, which were largely diluted in grammar and lexical details.
Concurrently, the general retain set is constructed by uniformly sampling broad-domain knowledge from Wikipedia and formulating it into a matching QA structure. 
For Obj2 (Neighboring Domain Retention), to accurately define and construct the neighboring domain set $\gN(\gD_f)$, we prompt an LLM to generate concepts that are semantically adjacent to but strictly outside the unlearning domain $\gD_f$ based on the published benchmark taxonomy. 
These extracted concepts are also standardized into QA pairs. 
For Obj3 (Adversarial Robustness), we approximate the inner maximization problem of adversarial prefilling: given a forgetting sample $x$, we employ the LLM (Gemini 3 Pro) to iteratively improve an adversarial prefix $x_p^{adv}$ that will maximally elicit the undesirable knowledge from the target model $\pi_\theta$.
%

\noindent \textbf{Data Standardization Tightens the Unlearning Bound:} 
Theoretically, by enforcing a consistent data representation, our method mitigates the structural discrepancies among unlearning domains, which also draws a connection to an improved generalization bound \citep{ben2006analysis}.
Specifically, consider the multi-objective unlearning problem as a multi-source domain
adaptation task, where each individual unlearning objective corresponds to a \textit{source} domain
$\mathcal{T}_k$, and the target domain $\mathcal{T}$ represents their ensemble, \textit{i.e.}, the
joint multi-objective unlearning goal. Let $h: \mathcal{X} \rightarrow \mathcal{Y}$ be a
hypothesis (model) over such domains, and $\mathcal{H} \subseteq \{h\}$ denote the hypothesis class. 
$d_{\mathcal{H}\Delta\mathcal{H}}
(\tilde{\mathcal{D}}_k, \tilde{\mathcal{D}})$ denotes the divergence measured over the
symmetric-difference hypothesis space between the $k$-th source domain $\tilde{\mathcal{D}}_k$
and the target domain $\tilde{\mathcal{D}}$. Then, with probability at least $1 - \delta$:
\begin{align} 
\begin{small}
\mathcal{L}_{\mathcal{T}}(h) \equiv \mathcal{L}_{\mathcal{T}}\left(\frac{1}{K}
\sum_{k} h_k\right) \leq \frac{1}{K}\sum_{k} \hat{\mathcal{L}}_{\mathcal{T}k}(h_k)
+ \frac{1}{K}\sum{k}\left(d_{\mathcal{H}\Delta\mathcal{H}}(\tilde{\mathcal{D}}_k,
\tilde{\mathcal{D}}) + \lambda_k\right) + C,
\end{small}
\end{align}
where $C$ is governed by the number of training samples and the
VC-dimension of $\mathcal{H}$. 
This bound reveals that the overall unlearning performance depends not only on the individual objective losses $\hat{\mathcal{L}}_{\mathcal{T}_k}
(h_k)$, but also on the distributional divergence between each source and the target domain.
In our setting, minimizing $d_{\mathcal{H}\Delta\mathcal{H}}(\tilde{\mathcal{D}}_k,
\tilde{\mathcal{D}})$ across unlearning domains directly tightens this bound, which aligns with the theoretical justification to serve as a principled path towards multi-objective unlearning optimization.

Empirically, this standardization ensures that the gradient updates for all seemingly isolated goals operate on a shared, harmonized geometric space. The impact of this data representation alignment is quantitatively validated in Table~\ref{fig:gradient_conflict}: the gradient updates of each unlearning goal shift dramatically to strong positive correlations (\textit{e.g.}, +92.29\% improvement between Obj1 and Obj2) after data standardization (the ``Ours'' column in Figure  \ref{fig:gradient_conflict}), which thus enables synergistic optimization across diverse unlearning objectives.

\vspace{-0.1in}
\subsection{Sharpening Unlearning Domain Boundaries via  Contrastive  Anchor Samples} \vspace{-0.1in}
\label{sec:contrastive_anchors}
%
While standardization aligns the optimization trajectory {across multiple unlearning goals}, the LLM still requires precise semantic boundaries to prevent the pervasive issue of \textit{over-refusal} on neighboring domains. 
{To resolve this ambiguity, we augment each objective's training set with \textit{contrastive anchor pairs}, each consisting of a positive example that the model should preserve or reject for the right reasons, and a negative example that sits near the decision boundary but belongs to a safe semantic territory. Concretely:}
%
    \textit{Obj1 Anchors}  contrast target hazardous knowledge against factually adjacent but benign world knowledge, so the model learns to distinguish what to erase from what to retain. 
    %
    \textit{Obj2 Anchors} are constructed between a restricted unlearning concept (\textit{\textit{e.g.,}} cyber-weaponry synthesis) and a semantically neighboring legitimate concept (\textit{\textit{e.g.,}} general computer security principles). 
    %
    \textit{Obj3 Anchors}  oppose a malicious prefilling prefix against a benign continuation, so the model develops robustness to adversarial elicitation without suppressing legitimate completions.

These contrastive pairs act analogously to  support vectors to help sculpt a more precise decision boundary during optimization, which defines the local geometry of the decision surface and prevents the forgetting signal from bleeding into adjacent safe regions.
%

\vspace{-0.1in}
\subsection{{Fine-Grained} Multi-Goal Unlearning via Bidirectional Top-K Logit Distillation} \vspace{-0.1in}
\label{sec:optimization_side}

Up to now, with the data representations aligned and boundaries anchored, we address the \textbf{optimization} side by proposing a \textit{Bidirectional Top-K Logit Distillation} mechanism. 
Consider $\pi^k_\text{ref} $ as a desirable reference model for each of the above unlearning tasks $\{\gT_k\}$, we reformat the proposed multi-goal unlearning problem as an imitation learning objective, 
\begin{align*}
\begin{small}
\min_\theta \E_{\gT_k \sim \{ \gT \}} \Big[ \E_{x\sim \gT_k} \Big[ \mathbb{D}[\pi_\theta(\cdot|x) \Vert \pi^k_\text{ref}(\cdot|x)]  \Big] \Big],
\end{small}
\end{align*}
%
where $\mathbb{D}[p \Vert q]$ can be a legitimate distance measure  between two probabilities of $p$ and $q$, such as the KL divergence or any $f$-divergence.

\textbf{Construction of reference model:} 
For each unlearning goal $k$, we utilize a lightweight natural language \textit{intention prompt}, $p_k$, which encodes the desired model behavior. A frozen teacher LLM, $\pi_{ref}$, is steered by $p_k$ and instructed via Chain of Thought (CoT) to establish safe and contextualized semantic boundaries. 
In this work, {we focus on conceptual unlearning, where the unlearning intention and scope can be naturally described by instructions, in contrast to entry-level unlearning~\citep{maini2024-tofu}.}
Then the student model, $\pi_{\theta}$, receives only the input $x$ without the intention prompt, forcing it to internalize the target behavior.

To achieve precise unlearning without distorting the global vocabulary distribution, we conduct distillation selectively on two critical sets of token logits: $\mathbb{C}_{K}^{teach}$ (the Top-$K$ logits from the teacher) and $\mathbb{C}_{K}^{stud}$ (the Top-$K$ logits spontaneously generated by the student).
\begin{small}
\begin{equation}
\mathbb{D}[\pi_\theta \Vert \pi_\text{ref}] \equiv L_{dual} = \underbrace{\mathbb{E}_{x} \left[ \sum_{i \in \mathbb{C}_{K}^{\text{ref}}} \mathcal{L}_{sim} \big( g_{\theta}^{i}(x), g_{\text{ref}}^{i}(x) \big) \right]}_{\text{Encouraging teacher imitation}} + \alpha \underbrace{\mathbb{E}_{x} \left[ \sum_{j \in \mathbb{C}_{K}^{\theta}} \mathcal{L}_{sim} \big( g_{\theta}^{j}(x), g_{\text{ref}}^{j}(x) \big) \right]}_{\text{Suppressing student's undesirable behavior}}
\end{equation}
\end{small}

where $g(\cdot)$ represents the logit outputs and $\mathcal{L}_{sim}$ denotes the Smooth L1 distance. This asymmetric mechanism operates bi-directionally:
1) It \textbf{\textit{encourages}} the student model toward the teacher's safe utility manifold by imitating the teacher's top probabilities, and meanwhile 2) explicitly \textbf{\textit{suppresses}} the student's high-confidence hazardous logits (\textit{i.e.}, the stubborn target knowledge), achieving precise memory erasure. Unlike prior distillation work that operates on the normalized logit probability space~\citep{hinton2015distilling,guminillm}, or uni-directional distillation~\citep{zhong2026duet}, our method achieves more effective, precise distillation effects (Section ~\ref{sec:ablation}).
Ultimately, this data-optimization framework enables the model to synergistically unlearn target data, maintain utility for neighboring data, and resist pre-filling attacks in a highly efficient manner.

%% file: sections/related_work.tex
\vspace{-0.1in}
\section{Related Work} \vspace{-0.1in}
\label{sec:related_work}

\paragraph{LLM Unlearning.}
Efforts to unlearn knowledge from LLMs can be broadly categorized into in-context methods \citep{pawelczyk2024incontextunlearninglanguagemodels,liu2024-eco-neurips}, which steer model behavior at inference time without parameter updates, and training-based methods \citep{jang2023-knowledge-unlearning-acl, zhang2024-npo, li2024-wmdp-icml, choi2024-snap, eldan2023whosharrypotterapproximate, maini2024tofutaskfictitiousunlearning, fan2025-simnpo, wang2025llm, pan2025multi, jin2025unlearning}, which modify model weights to enforce forgetting. 
Recently, an emerging line of work has investigated the critical impact of \textbf{\textit{data}} design and selection on unlearning performance \citep{new_data_unlearning_1, new_data_unlearning_2}. 
However, methods in this paradigm mostly focus on a setting captured by our \textbf{Obj1}, \textit{i.e.,} a dual-objective setting comprising a forgetting domain and an irrelevant general domain. They largely overlook the complex boundary behaviors and adversarial vulnerabilities inherent in real-world deployments.

\vspace{-0.1in}
\paragraph{Unlearning Robustness.}
A critical dimension of unlearning is the robustness of the ``forgotten'' state. Prior works have highlighted vulnerabilities against parameterized attacks, such as relearning attacks where fine-tuning on a handful of examples restores the erased capabilities \citep{fan2025towards,hu2024-jogging-memory,qifine}. 
Furthermore, recent studies demonstrate that unlearned knowledge can still be extracted via contextualized attacks, which perfectly aligns with our setting. For instance, \cite{shumailov2024ununlearningunlearningsufficientcontent} and \cite{lucki2025adversarialperspectivemachineunlearning} formalized in-context un-unlearning and adversarial jailbreaks, where sophisticated contextual elicitation strategies can bypass unlearning guardrails \citep{liu2024contextual}. 
%


\vspace{-0.1in}
\paragraph{LLM Knowledge Distillation.} 
Knowledge distillation (KD) has been extensively leveraged to transfer capabilities across LLMs \citep{li2025bild, guminillm}. 
In the context of unlearning, while KD is commonly employed as a regularization penalty on the retain set to mitigate catastrophic forgetting \citep{zhang2024-npo, jang2023-knowledge-unlearning-acl, yao2024large}, utilizing distillation as the primary mechanism to directly enforce forgetting remains highly underexplored.
Specifically, \citep{zhong2026duet} proposed DUET, a uni-directional unlearning distillation method for aligning a student model with a prompt-steered teacher. 
Building upon this foundation, our work extends distillation into a bidirectional framework with richer and more explicit supervision signals.

\vspace{-0.1in}
\paragraph{LLM Unlearning Benchmarks.} 
Evaluation remains a fundamental challenge in LLM unlearning. Early benchmarks such as TOFU \citep{maini2024tofutaskfictitiousunlearning} and MUSE \citep{shi2024muse} focused primarily on exact memorization and general utility. 
While more recent benchmarks like WMDP \citep{li2024-wmdp-icml} and RWKU \citep{jin2024rwku} introduce high-risk capabilities and partially discuss unlearning robustness, they lack a systematic investigation into boundary behaviors, conceptual unlearning conflicts, and the collateral damage to neighboring domains. 
Our experimental setup bridges this gap by augmenting existing benchmarks with structured evaluations (Obj1--Obj3) that capture the holistic demands of practical unlearning: efficacy, neighboring domain retention, and adversarial defense.

%% file: sections/experiments.tex
\vspace{-0.1in}
\section{Experiments} \vspace{-0.15in}
\label{sec:experiments}

%
In this section, we empirically evaluate how the proposed underlying representational alignment and assymetric bidirectional unlearning distillation translate into improved unlearning performance relative to state-of-the-art unlearning methods.
We also provide a fine-grained ablation study to validate the necessity of our designed components.

\vspace{-0.1in}
\subsection{Experimental Setup} \vspace{-0.1in}
\label{sec:exp_setup}

\paragraph{Dataset Construction and Metrics.}
We evaluated our framework across two distinct knowledge domains: the \textbf{MUSE-Book} benchmark (\textit{i.e.,} {Harry Potter (HP)}) ~\citep{shi2024-muse}, which represents copyrighted narrative knowledge, and the \textbf{WMDP-Cyber} benchmark~\citep{li2024wmdp}, which represents hazardous, restricted knowledge. 
%
%
We applied the following evaluation metrics to reflect our three unlearning objectives, and additionally reported the \textbf{MMLU} scores~\citep{hendrycks2020measuring} to monitor the unlearned model's general capability.

\begin{itemize}[leftmargin=*] \vspace{-0.1in}
    \item \textbf{Obj1} (Target Unlearning \& General Utility):  We measured the unlearning efficacy on target knowledge (\textbf{Forget} $\downarrow$) and preservation on general knowledge (\textbf{Retain} $\uparrow$), evaluated on open-ended questions, using the \textit{ROUGE scores} against reference targets as the metric;
    \item \textbf{Obj2} (Neighboring Domain Retention):   We measured the model's ability to preserve knowledge on this benign neighboring domain while  avoiding \textit{over-refusal}.
    In particular, to obtain a fine-grained evaluation of the model's performance on this task, we used a Multi-Choice Question (MCQ) format for evaluation, where each query includes plausible yet confusing options, refusal options (\textit{e.g.,} ``I do not know''), and the ground-truth answer option. We adopted the \textbf{Retain Acc} $\uparrow$ as the main metric.
    \item \textbf{Obj3} (Robustness Against Adversarial Prompts): We prefilled each query related to the unlearning domain with an adversarial prefilling prompt.  We adopted the Attack Success Rate (\textbf{ASR} $\downarrow$) to indicate the rate at which the target model is manipulated with malicious prefixes. Evaluation queries for this task are drawn from the same unlearning domain but are disjoint from those in Obj1.
\end{itemize}

\paragraph{Unlearning Baselines and Upper Bound.}
We compare our method against three categories of approaches to clearly distinguish between different unlearning paradigms: \vspace{-0.1in}
\begin{enumerate}[leftmargin=*]
    \item  \textbf{In Context Unlearning Methods}:
    \begin{itemize}[leftmargin=*]
    \item \textit{Teacher with Oracle Prompt Router} ({Upper Bound}): An idealized configuration under optimal, non-conflicting conditions. We approximate this by assuming a perfect oracle router that assigns the exact goal-specific intention prompt ($p_m$) to steer the base model for each corresponding query.
     \item \textit{Ensemble Teacher},  which concatenates all intention prompts into a single instruction to steer the base model, acting as a practical alternative to the oracle router.
     \end{itemize}
    
    \item \textbf{Training Based Unlearning Methods:}
        \textit{ DUET (Distilled from the Ensemble Teacher):} An unlearned model trained using the distillation method in DUET~\citep{zhong2026duet} to imitate the Ensemble Teacher. 
        \textit{Supervised Fine-Tuning:} A standard training approach that maximizes the generation of desirable responses~\citep{touvron2023llama2openfoundation}, including producing refusal responses for forget or adversarial queries, and accurate answers for retention and neighboring queries. It serves as a fundamental baseline for comparison.
        \textit{Gradient Ascent (GA)} \citep{jang2023-knowledge-unlearning-acl}, \textit{SimNPO} \citep{fan2025-simnpo}, and \textit{FLAT} \citep{wang2025-flat-iclr}: unlearning methods primarily proposed for addressing Obj1.  
    
    \item \textbf{Multi-Objective Unlearning with Gradient Editing:} These two methods below were originally designed to address the forget-retention tradeoff (Obj1). We extended their methods to handle three objectives. Especially:
    \textit{NGDiff}~\citep{jin2025unlearning} addresses unlearning objective conflicts by utilizing normalized gradient differences and an adaptive learning rate; 
 \textit{MOLLM}~\citep{pan2025multi} employs gradient projection mechanisms to explicitly balance different unlearning objectives.
\end{enumerate}

\vspace{-0.1in}
\subsection{Main Results} 
\label{sec:main_results}\vspace{-0.1in}

\begin{wrapfigure}[15]{r}{0.5\textwidth}  
    \centering
    \vspace{-0.2in} 
    \includegraphics[width=0.48\textwidth]{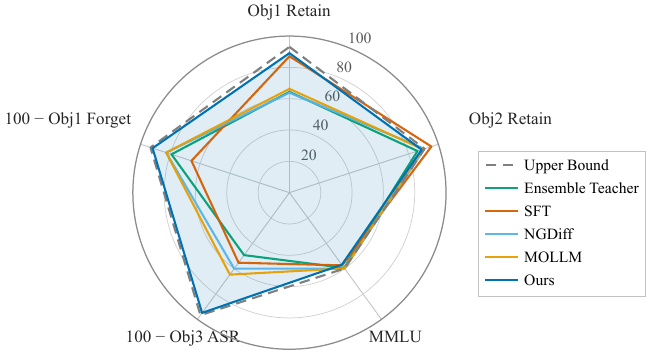}\vspace{-0.15in}
    \caption{\small{Multi-dimensional performance on the \textbf{WMDP-Cyber} benchmark. The radar chart illustrates the trade-offs across five metrics axes (Forget $\downarrow$, Retain $\uparrow$,  Neighbor Retain Acc $\uparrow$, ASR $\downarrow$, MMLU $\uparrow$),  normalized for visualization.} }
    \label{fig:wmdp_radar}
     \vspace{-0.2in} 
\end{wrapfigure}

The main results on the MUSE-Book (Harry Potter) benchmark are reported in Table \ref{tab:main_hp}. To provide a comprehensive view of multi-dimensional trade-offs on the WMDP-Cyber benchmark, we visualized performance across all objectives using a radar chart in Figure \ref{fig:wmdp_radar}. These two main results show that our proposed framework effectively achieves multi-goal unlearning while mitigating the performance imbalances observed in prior methods.

\vspace{-0.05in}

\begin{table*}[h]
\centering

\resizebox{\textwidth}{!}{
\begin{tabular}{lcccccc}
\toprule
\multirow{2}{*}{\textbf{Method}} 
& \multicolumn{2}{c}{\textbf{Obj1} (Unlearn)} 
& \textbf{Obj2} (Neighbor Domain)  
& \textbf{Obj3} (Adv Robustness) 
& \multirow{2}{*}{\textbf{MMLU $\uparrow$}} 
& \multirow{2}{*}{\makecell{\textbf{Overall}\\\textbf{Performance $\uparrow$}}} \\
\cmidrule(lr){2-3}
\cmidrule(lr){4-4}
\cmidrule(lr){5-5}
& \textbf{Forget  $\downarrow$} 
& \textbf{General  $\uparrow$} 
& \textbf{Retain Acc $\uparrow$} 
& \textbf{ASR $\downarrow$} 
&  &  \\
\midrule
Base Model      & 32.1 & 84.3 & 57.28 & 70.5 & \textbf{60.8} & 0.00 \\
\addlinespace[3pt]
\midrule
\addlinespace[3pt]

GA ($\star$)             & 36.9 & \textbf{85.0} & 55.9 & 70.0 & 36.45 & -29.33 \\
SimNPO ($\star$)         & 21.4 & 43.1 & 57.7 & 70.5 & 60.40 & -30.48 \\
FLAT ($\star$)           & 0.7  & 58.3 & 42.7 & 39.0 & 58.92 & 20.44 \\ \midrule

SFT                        & \textbf{0.3} & 49.6 & 50.7 & \textbf{1.5} & 50.2 & 48.92 \\
NGDiff                     & 13.6 & 83.3 & 37.7 & 17.9 & \textbf{60.8} & 50.52 \\
MOLLM                      & 19.9 & 83.6 & 57.4 & 69.2 & \textbf{60.8} & 12.92 \\
Ensemble Teacher (ET)           & 6.2  & 60.9 & 26.8 & 13.5 & 60.3 & 28.52 \\
DUET (w/ ET) & 3.2  & 71.3 & 26.8 & 39.5 & 49.9 & 5.52 \\

\textbf{Ours}              & 2.7 & 78.1 & \textbf{58.5} & 12.5 & 59.2 & \textbf{80.82} \\
\addlinespace[2pt]

\midrule
\textit{Upper Bound}
                           & \textit{2.1} & \textit{80.1} & \textit{58.5} & \textit{6.7} & \textit{60.7} & \textit{90.72} \\
\bottomrule
\end{tabular}
}
\vspace{-0.5em}
\caption{\small{Main unlearning results on the \textbf{MUSE-Book} benchmark. We also reported the aggregated \textbf{Overall Performance} shift, defined relative to the base model. 
{Methods marked with '$\star$' are trained on Obj1 due to their inherent design and are not readily extendable to multi-goal settings.}}}
\label{tab:main_hp}

\end{table*}

\textbf{Shortcomings of Naive Baselines.} As observed in Table \ref{tab:main_hp}, while the Ensemble Teacher achieves moderate forgetting, it suffers a significant drop in retention compared to the Upper Bound. Since both configurations employ in-context unlearning, this degradation clearly indicates that concatenating multiple intention prompts leads to internal intention interference; the LLM struggles to satisfy these instructions simultaneously. 
Furthermore, even after applying standard distillation, \textit{DUET} exhibited persistent trade-offs with a notable degradation in   Obj3 ASR (39.5\%) and MMLU (49.9\%). 
Finally, SFT aggressively destroys the target knowledge (Obj1 Forget drops to 0.3) while effectively defending against prefilling attacks (ASR drops to 1.5\%). However, this result comes at the cost of catastrophic \textit{over-refusal}, with  degraded general utility (Obj1 and Obj2 Retain collapse to 49.6 and 50.7, respectively, and MMLU plummets to 50.2).

\textbf{Imbalanced Trade-offs in Gradient Editing Methods.} While NGDiff and MOLLM address multi-objective conflicts through gradient scalarization and orthogonal projection without data standardization, they exhibit weakened unlearning robustness.
On the HP benchmark, MOLLM remains highly vulnerable to prefilling attacks (ASR $69.2\%$) and shows limited effectiveness in unlearning QA targets (Obj1 Forget $19.9$). Similarly, NGDiff notably underperforms on both ASR ($17.9\%$) and target forgetting (Obj1 Forget $13.6$). Although both methods retain strong performance on Retain and MMLU, this is largely because they fail to sufficiently remove hazardous representations.

In contrast, our method achieves strong synergy across objectives. By unifying data representations and leveraging intention-prompted bidirectional distillation, we substantially reduce prefilling ASR to $12.5\%$, outperforming gradient editing-based methods by an immense margin while effectively erasing target knowledge (Obj1 Forget $2.7$). Simultaneously, our method strictly prevents over-refusal, maintaining high retention accuracy on general and neighboring domains (Obj1 Retain $78.1\%$, Obj2 Retain Acc $58.9\%$), which  matches the Upper Bound performance.


\textbf{Performance on WMDP-Cyber.} The results in Figure \ref{fig:wmdp_radar} reinforce the above findings in the hazardous knowledge domain. The Upper Bound forms a large, well-balanced polygon that represents ideal performance across all objectives. Baselines such as SFT achieve strong security but exhibit a pronounced collapse along the retention axes. Conversely, methods like MOLLM and NGDiff preserve utility but degrade substantially on ASR and unlearning objectives.
Our framework is the only approach that maintains a balanced performance profile, effectively mitigating hazardous knowledge and adversarial attacks while preserving the integrity of neighboring-domain retention.

 \vspace{-0.1in}
\subsection{Ablation Study} \vspace{-0.1in}
\label{sec:ablation}


\begin{wraptable}[11]{r}{0.5\textwidth}
\vspace{-1.25em}
\centering
\tiny
\scriptsize
\setlength{\tabcolsep}{2pt}
\begin{tabular}{p{1.44cm}cccccc}
\toprule
\multirow{2}{*}{\textbf{Method}} 
& \multicolumn{2}{c}{\textbf{Obj1}} 
& \textbf{Obj2} 
& \textbf{Obj3} 
& \multirow{2}{*}{\textbf{MMLU}} 
& \multirow{2}{*}{\shortstack{\textbf{Overall}\\\textbf{Perf.}}} \\
\cmidrule(lr){2-3}
\cmidrule(lr){4-4}
\cmidrule(lr){5-5}
& \textbf{Forget} 
& \textbf{General} 
& \textbf{Retain} 
& \textbf{ASR} 
&  &  \\
\midrule
\textbf{Ours} 
& \textbf{2.7} & 78.1 
& \textbf{58.5} 
& \textbf{12.5} 
& 58.2 
& \textbf{79.82} \\
\addlinespace[2pt]

Ours w/ \newline \textbf{DUET} Loss
& 3.19 & 74.0
& 57.6 
& \textbf{12.5} 
& 58.2 
& 74.33 \\
\addlinespace[2pt]

Ours w/ \newline \textbf{BILD} Loss
& 3.9 & \textbf{82.3} 
& 56.3
& 23.5 
& \textbf{58.3} 
& 69.72 \\
\bottomrule
\end{tabular}
\vspace{-1em}
\caption{
Ablation study replacing our asymmetric bidirectional distillation with a symmetric BILD objective.
}
\label{tab:ablation_bild}
\end{wraptable}

To evaluate the effects of our {bidirectional top-K logit distillation}, we conducted a comparative analysis against a state-of-the-art symmetric distillation approach, the bi-directional Logits Difference (BILD) loss \citep{li2025bild}. 
BILD constructs pairwise logit differences within an active vocabulary and \textit{\textbf{normalizes}} them into probability distributions via a Softmax function, then minimizes their KL divergence:
$ \mathcal{L}_{BiLD} \propto D_{KL}(p^t \parallel p^s) + D_{KL}(p^s \parallel p^t),$ 
where $p^t$ and $p^s$ denote the Softmax-normalized probability distributions of the teacher and student top-K logits, respectively. 
While effective for standard capability transfer, this \textit{normalized KL} approach fundamentally only aligns the relative distances (rankings) between tokens. 
In contrast, our method explicitly minimizes differences in magnitude using a \textit{Smooth L1} distance applied directly tostance directly on the unnormalized logits.





As demonstrated in Table \ref{tab:ablation_bild}, replacing our method with BILD leads to a significant regression in unlearning efficacy (Obj1 Forget rises to $3.9$) and adversarial robustness (ASR almost doubles to $23.5\%$). 
This ablation empirically proves our  claim that, in the context of unlearning, aligning only the \textit{relative ranking} of logits is insufficient.
The performance gains over DUET with uni-directional distillation further indicate that effectively unlearning targeted knowledge requires our asymmetric distillation strategy, which explicitly suppresses the student’s high-confidence  logits toward the teacher's logits.

%% file: colm2026_conference.bib
@misc{li2024wmdpbenchmarkmeasuringreducing,
      title={The WMDP Benchmark: Measuring and Reducing Malicious Use With Unlearning}, 
      author={Nathaniel Li and Alexander Pan and Anjali Gopal and Summer Yue and Daniel Berrios and Alice Gatti and Justin D. Li and Ann-Kathrin Dombrowski and Shashwat Goel and Long Phan and Gabriel Mukobi and Nathan Helm-Burger and Rassin Lababidi and Lennart Justen and Andrew B. Liu and Michael Chen and Isabelle Barrass and Oliver Zhang and Xiaoyuan Zhu and Rishub Tamirisa and Bhrugu Bharathi and Adam Khoja and Zhenqi Zhao and Ariel Herbert-Voss and Cort B. Breuer and Samuel Marks and Oam Patel and Andy Zou and Mantas Mazeika and Zifan Wang and Palash Oswal and Weiran Lin and Adam A. Hunt and Justin Tienken-Harder and Kevin Y. Shih and Kemper Talley and John Guan and Russell Kaplan and Ian Steneker and David Campbell and Brad Jokubaitis and Alex Levinson and Jean Wang and William Qian and Kallol Krishna Karmakar and Steven Basart and Stephen Fitz and Mindy Levine and Ponnurangam Kumaraguru and Uday Tupakula and Vijay Varadharajan and Ruoyu Wang and Yan Shoshitaishvili and Jimmy Ba and Kevin M. Esvelt and Alexandr Wang and Dan Hendrycks},
      year={2024},
      eprint={2403.03218},
      archivePrefix={arXiv},
      primaryClass={cs.LG},
      url={https://arxiv.org/abs/2403.03218}, 
}

@article{li2024wmdp,
  title={The wmdp benchmark: Measuring and reducing malicious use with unlearning},
  author={Li, Nathaniel and Pan, Alexander and Gopal, Anjali and Yue, Summer and Berrios, Daniel and Gatti, Alice and Li, Justin D and Dombrowski, Ann-Kathrin and Goel, Shashwat and Phan, Long and others},
  journal={Proceedings of Machine Learning Research},
  year={2024}
}

@article{pawelczyk2023context,
  title={In-context unlearning: Language models as few shot unlearners},
  author={Pawelczyk, Martin and Neel, Seth and Lakkaraju, Himabindu},
  journal={arXiv preprint arXiv:2310.07579},
  year={2023}
}

@article{yao2024large,
  title={Large language model unlearning},
  author={Yao, Yuanshun and Xu, Xiaojun and Liu, Yang},
  journal={Advances in Neural Information Processing Systems},
  volume={37},
  pages={105425--105475},
  year={2024}
}

@misc{eldan2023whosharrypotterapproximate,
      title={Who's Harry Potter? Approximate Unlearning in LLMs}, 
      author={Ronen Eldan and Mark Russinovich},
      year={2023},
      eprint={2310.02238},
      archivePrefix={arXiv},
      primaryClass={cs.CL},
      url={https://arxiv.org/abs/2310.02238}, 
}

@article{maini2024tofutaskfictitiousunlearning,
  title={TOFU: A Task of Fictitious Unlearning for LLMs},
  author={Pratyush Maini and Zhili Feng and Avi Schwarzschild and Zachary C. Lipton and J. Zico Kolter},
  journal={First Conference on Language Modeling},
  year={2024}
}

@misc{shumailov2024ununlearningunlearningsufficientcontent,
      title={UnUnlearning: Unlearning is not sufficient for content regulation in advanced generative AI}, 
      author={Ilia Shumailov and Jamie Hayes and Eleni Triantafillou and Guillermo Ortiz-Jimenez and Nicolas Papernot and Matthew Jagielski and Itay Yona and Heidi Howard and Eugene Bagdasaryan},
      year={2024},
      eprint={2407.00106},
      archivePrefix={arXiv},
      primaryClass={cs.LG},
      url={https://arxiv.org/abs/2407.00106}, 
}

@misc{pawelczyk2024incontextunlearninglanguagemodels,
      title={In-Context Unlearning: Language Models as Few Shot Unlearners}, 
      author={Martin Pawelczyk and Seth Neel and Himabindu Lakkaraju},
      year={2024},
      eprint={2310.07579},
      archivePrefix={arXiv},
      primaryClass={cs.LG},
      url={https://arxiv.org/abs/2310.07579}, 
}

@inproceedings{jang2023-knowledge-unlearning-acl,
  title     = {Knowledge Unlearning for Mitigating Privacy Risks in Language Models},
  author    = {Jang, Joel and Yoon, Dongkeun and Yang, Sohee and Cha, Sungmin and Lee, Moontae and Logeswaran, Lajanugen and Seo, Minjoon},
  booktitle = {Proceedings of the 61st Annual Meeting of the Association for Computational Linguistics (ACL)},
  year      = {2023},
  pages     = {14389--14408},
  url       = {https://aclanthology.org/2023.acl-long.805}
}

@inproceedings{liu2024-eco-neurips,
  title     = {Large Language Model Unlearning via Embedding-Corrupted Prompts},
  author    = {Liu, Chris and Li, Zeyu and Wei, Yujia and Wang, Peizhuo and Li, Chen and Zhang, Wei and Li, Song and Magdon-Ismail, Malik and Liu, Yang},
  booktitle = {Advances in Neural Information Processing Systems (NeurIPS)},
  year      = {2024},
  url       = {https://arxiv.org/abs/2406.07933}
}

@article{hu2024-jogging-memory,
  title   = {Jogging the Memory of Unlearned LLMs Through Targeted Relearning Attacks},
  author  = {Hu, Shengyuan and Fu, Yiwei and Wu, Zhiwei Steven and Smith, Virginia},
  journal = {arXiv preprint arXiv:2406.13356},
  year    = {2024},
  url     = {https://arxiv.org/abs/2406.13356}
}

@article{zhang2024-npo,
  title   = {Negative Preference Optimization for Catastrophic Forgetting in LLM Unlearning},
  author  = {Zhang, Yuhan and Chen, Weize and others},
  journal = {arXiv preprint arXiv:2404.05868},
  year    = {2024},
  url     = {https://arxiv.org/abs/2404.05868}
}

@article{fan2025-simnpo,
  title   = {Simplicity Prevails: Rethinking Negative Preference Optimization for LLM Unlearning},
  author  = {Fan, Chongyu and Liu, Jiancheng and Lin, Licong and Jia, Jinghan and Zhang, Ruiqi and Mei, Song and Liu, Sijia},
  journal = {arXiv preprint arXiv:2410.07163},
  year    = {2025},
  url     = {https://arxiv.org/abs/2410.07163}
}

@article{choi2024-snap,
  title   = {SNAP: Unlearning Selective Knowledge in Large Language Models with Negative Instructions},
  author  = {Choi, Minseok and Rim, Daniel and Lee, Dohyun and Choo, Jaegul},
  journal = {arXiv preprint arXiv:2406.12329},
  year    = {2024},
  url     = {https://arxiv.org/abs/2406.12329}
}

@article{maini2024-tofu,
  title   = {TOFU: A Task of Fictitious Unlearning for LLMs},
  author  = {Maini, Pratyush and Jain, Shrey and others},
  journal = {arXiv preprint arXiv:2401.06121},
  year    = {2024},
  url     = {https://arxiv.org/abs/2401.06121}
}

@article{shi2024-muse,
  title   = {MUSE: Machine Unlearning Six-Way Evaluation for Language Models},
  author  = {Shi, Weijia and Holtzman, Ari and Raffel, Colin and others},
  journal = {arXiv preprint arXiv:2407.06460},
  year    = {2024},
  url     = {https://arxiv.org/abs/2407.06460}
}

@inproceedings{li2024-wmdp-icml,
  title     = {The WMDP Benchmark: Measuring and Reducing Malicious Use with Unlearning},
  author    = {Li, Nathaniel and Pan, Alexander and Gopal, Anjali and Yue, Summer and Berrios, Daniel and Gatti, Alice and Li, Justin D. and Dombrowski, Ann-Kathrin and Goel, Shashwat and Mukobi, Gabriel and others},
  booktitle = {Proceedings of the 41st International Conference on Machine Learning (ICML)},
  series    = {Proceedings of Machine Learning Research},
  volume    = {235},
  pages     = {28525--28550},
  year      = {2024},
  url       = {https://proceedings.mlr.press/v235/li24bc.html}
}

@inproceedings{li2025bild,
  title={Bild: Bi-directional logits difference loss for large language model distillation},
  author={Li, Minchong and Zhou, Feng and Song, Xiaohui},
  booktitle={Proceedings of the 31st International Conference on Computational Linguistics},
  pages={1168--1182},
  year={2025}
}

@inproceedings{kifer2004detecting,
  title={Detecting change in data streams},
  author={Kifer, Daniel and Ben-David, Shai and Gehrke, Johannes},
  booktitle={VLDB},
  volume={4},
  pages={180--191},
  year={2004},
  organization={Toronto, Canada}
}

@article{ben2006analysis,
  title={Analysis of representations for domain adaptation},
  author={Ben-David, Shai and Blitzer, John and Crammer, Koby and Pereira, Fernando},
  journal={Advances in neural information processing systems},
  volume={19},
  year={2006}
}

@misc{zhong2025hierarchicalfederatedunlearninglarge,
      title={Hierarchical Federated Unlearning for Large Language Models}, 
      author={Yisheng Zhong and Zhengbang Yang and Zhuangdi Zhu},
      year={2025},
      eprint={2510.17895},
      archivePrefix={arXiv},
      primaryClass={cs.LG},
      url={https://arxiv.org/abs/2510.17895}, 
}

@article{nguyen2024surveymachineunlearning,
title={A survey of machine unlearning},
author={Nguyen, Thanh Tam and Huynh, Thanh Trung and Ren, Zhao and Nguyen, Phi Le and Liew, Alan Wee-Chung and Yin, Hongzhi and Nguyen, Quoc Viet Hung},
journal={ACM Transactions on Intelligent Systems and Technology},
volume={16},
number={5},
pages={1--46},
year={2025},
publisher={ACM New York, NY}
}

@article{xu2023machineunlearningsurvey,
author = {Xu, Heng and Zhu, Tianqing and Zhang, Lefeng and Zhou, Wanlei and Yu, Philip S.},
title = {Machine Unlearning: A Survey},
year = {2023},
issue_date = {January 2024},
publisher = {Association for Computing Machinery},
address = {New York, NY, USA},
volume = {56},
number = {1},
issn = {0360-0300},
url = {https://doi.org/10.1145/3603620},
doi = {10.1145/3603620},
journal = {ACM Comput. Surv.},
month = aug,
articleno = {9},
numpages = {36}
}

@inproceedings{pawelczykcontext,
  title={In-Context Unlearning: Language Models as Few-Shot Unlearners},
  author={Pawelczyk, Martin and Neel, Seth and Lakkaraju, Himabindu},
  booktitle={Forty-first International Conference on Machine Learning}
}

@inproceedings{takashiro-etal-2025-answer,
    title = "Answer When Needed, Forget When Not: Language Models Pretend to Forget via In-Context Knowledge Unlearning",
    author = "Takashiro, Shota  and
      Kojima, Takeshi  and
      Gambardella, Andrew  and
      Cao, Qi  and
      Iwasawa, Yusuke  and
      Matsuo, Yutaka",
    editor = "Che, Wanxiang  and
      Nabende, Joyce  and
      Shutova, Ekaterina  and
      Pilehvar, Mohammad Taher",
    booktitle = "Findings of the Association for Computational Linguistics: ACL 2025",
    month = jul,
    year = "2025",
    address = "Vienna, Austria",
    publisher = "Association for Computational Linguistics",
    url = "https://aclanthology.org/2025.findings-acl.1276/",
    doi = "10.18653/v1/2025.findings-acl.1276",
    pages = "24872--24885",
    ISBN = "979-8-89176-256-5"
}

@InProceedings{pmlr-v267-muresanu25a,
  title = 	 {Fast Exact Unlearning for In-Context Learning Data for {LLM}s},
  author =       {Muresanu, Andrei Ioan and Thudi, Anvith and Zhang, Michael R. and Papernot, Nicolas},
  booktitle = 	 {Proceedings of the 42nd International Conference on Machine Learning},
  pages = 	 {45272--45288},
  year = 	 {2025},
  editor = 	 {Singh, Aarti and Fazel, Maryam and Hsu, Daniel and Lacoste-Julien, Simon and Berkenkamp, Felix and Maharaj, Tegan and Wagstaff, Kiri and Zhu, Jerry},
  volume = 	 {267},
  series = 	 {Proceedings of Machine Learning Research},
  month = 	 {13--19 Jul},
  publisher =    {PMLR},
  url = 	 {https://proceedings.mlr.press/v267/muresanu25a.html}
}

@inproceedings{jin2025unlearning,
  title={Unlearning as multi-task optimization: A normalized gradient difference approach with an adaptive learning rate},
  author={Jin, Xiaomeng and Bu, Zhiqi and Vinzamuri, Bhanukiran and Ramakrishna, Anil and Chang, Kai-Wei and Cevher, Volkan and Hong, Mingyi},
  booktitle={Proceedings of the 2025 Conference of the Nations of the Americas Chapter of the Association for Computational Linguistics: Human Language Technologies (Volume 1: Long Papers)},
  pages={11278--11294},
  year={2025}
}

@inproceedings{pan2025multi,
  title={Multi-Objective Large Language Model Unlearning},
  author={Pan, Zibin and Zhang, Shuwen and Zheng, Yuesheng and Li, Chi and Cheng, Yuheng and Zhao, Junhua},
  booktitle={ICASSP 2025-2025 IEEE International Conference on Acoustics, Speech and Signal Processing (ICASSP)},
  pages={1--5},
  year={2025},
  organization={IEEE}
}

@article{hendrycks2020measuring,
  title={Measuring massive multitask language understanding},
  author={Hendrycks, Dan and Burns, Collin and Basart, Steven and Zou, Andy and Mazeika, Mantas and Song, Dawn and Steinhardt, Jacob},
  journal={International Conference on Learning Representations},
  year={2020}
}

@article{jin2024rwku,
  title={Rwku: Benchmarking real-world knowledge unlearning for large language models},
  author={Jin, Zhuoran and Cao, Pengfei and Wang, Chenhao and He, Zhitao and Yuan, Hongbang and Li, Jiachun and Chen, Yubo and Liu, Kang and Zhao, Jun},
  journal={The Thirty-eight Conference on Neural Information Processing Systems Datasets and Benchmarks Track
},
  year={2024}
}

@article{shi2024muse,
  title={Muse: Machine unlearning six-way evaluation for language models},
  author={Shi, Weijia and Lee, Jaechan and Huang, Yangsibo and Malladi, Sadhika and Zhao, Jieyu and Holtzman, Ari and Liu, Daogao and Zettlemoyer, Luke and Smith, Noah A and Zhang, Chiyuan},
  journal={The Thirteenth International Conference on Learning Representations},
  year={2024}
}

@article{zhang2024negative,
  title={Negative preference optimization: From catastrophic collapse to effective unlearning},
  author={Zhang, Ruiqi and Lin, Licong and Bai, Yu and Mei, Song},
  journal={Conference on Language Modeling},
  year={2024}
}

@article{wang2025llm,
  title={{LLM} Unlearning via Loss Adjustment with Only Forget Data},
  author={Wang, Yaxuan and Wei, Jiaheng and Liu, Chris Yuhao and Pang, Jinlong and Liu, Quan and Shah, Ankit Parag and Bao, Yujia and Liu, Yang and Wei, Wei},
  journal={The Thirteenth International Conference on Learning Representations},
  year={2025}
}

@InProceedings{pmlr-v132-neel21a,
  title = 	 {Descent-to-Delete: Gradient-Based Methods for Machine Unlearning},
  author =       {Neel, Seth and Roth, Aaron and Sharifi-Malvajerdi, Saeed},
  booktitle = 	 {Proceedings of the 32nd International Conference on Algorithmic Learning Theory},
  pages = 	 {931--962},
  year = 	 {2021},
  editor = 	 {Feldman, Vitaly and Ligett, Katrina and Sabato, Sivan},
  volume = 	 {132},
  series = 	 {Proceedings of Machine Learning Research},
  month = 	 {16--19 Mar},
  publisher =    {PMLR},
  url = 	 {https://proceedings.mlr.press/v132/neel21a.html}
}

@article{lucki2025adversarialperspectivemachineunlearning,
  title={An adversarial perspective on machine unlearning for ai safety},
  author={{\L}ucki, Jakub and Wei, Boyi and Huang, Yangsibo and Henderson, Peter and Tram{\`e}r, Florian and Rando, Javier},
  journal={arXiv preprint arXiv:2409.18025},
  year={2024}
}

@inproceedings{wang2025-flat-iclr,
  title     = {{LLM} Unlearning via Loss Adjustment with Only Forget Data},
  author    = {Wang, Yaxuan and Wei, Jiaheng and Liu, Chris Yuhao and Pang, Jinlong and Liu, Quan and Shah, Ankit Parag and Bao, Yujia and Liu, Yang and Wei, Wei},
  booktitle = {International Conference on Learning Representations (ICLR)},
  year      = {2025},
  url       = {https://proceedings.iclr.cc/paper_files/paper/2025/file/6b315c0b736711b56f33cbacfb6d5d67-Paper-Conference.pdf}
}

@inproceedings{new_data_unlearning_1,
  title={Which retain set matters for llm unlearning? a case study on entity unlearning},
  author={Chang, Hwan and Lee, Hwanhee},
  booktitle={Findings of the Association for Computational Linguistics: ACL 2025},
  pages={5966--5982},
  year={2025}
}

@article{new_data_unlearning_2,
  title={From Domains to Instances: Dual-Granularity Data Synthesis for LLM Unlearning},
  author={Xu, Xiaoyu and Du, Minxin and Li, Zitong and Liang, Zi and Guo, Zhibiao and Zhang, Shiyu and Hu, Peizhao and Ye, Qingqing and Hu, Haibo},
  journal={arXiv preprint arXiv:2601.04278},
  year={2026}
}

@article{liu2024contextual,
  title={No Free Lunch for Defending Against Prefilling Attack by In-Context Learning},
  author={Xue, Zhiyu and Liu, Guangliang and Chen, Bocheng and Johnson, Kristen Marie and Pedarsani, Ramtin},
  journal={arXiv preprint arXiv:2412.12192},
  year={2024}
}

@article{zhong2026duet,
  title={DUET: Distilled LLM Unlearning from an Efficiently Contextualized Teacher},
  author={Zhong, Yisheng and Yang, Zhengbang and Zhu, Zhuangdi},
  journal={ICLR},
  year={2026}
}

@article{andriushchenko2024jailbreaking,
  title={Jailbreaking leading safety-aligned llms with simple adaptive attacks},
  author={Andriushchenko, Maksym and Croce, Francesco and Flammarion, Nicolas},
  journal={arXiv preprint arXiv:2404.02151},
  year={2024}
}

@article{fan2025towards,
  title={Towards llm unlearning resilient to relearning attacks: A sharpness-aware minimization perspective and beyond},
  author={Fan, Chongyu and Jia, Jinghan and Zhang, Yihua and Ramakrishna, Anil and Hong, Mingyi and Liu, Sijia},
  journal={ICML},
  year={2025}
}

@misc{touvron2023llama2openfoundation,
      title={Llama 2: Open Foundation and Fine-Tuned Chat Models}, 
      author={Hugo Touvron and Louis Martin and Kevin Stone and Peter Albert and Amjad Almahairi and Yasmine Babaei and Nikolay Bashlykov and Soumya Batra and Prajjwal Bhargava and Shruti Bhosale and Dan Bikel and Lukas Blecher and Cristian Canton Ferrer and Moya Chen and Guillem Cucurull and David Esiobu and Jude Fernandes and Jeremy Fu and Wenyin Fu and Brian Fuller and Cynthia Gao and Vedanuj Goswami and Naman Goyal and Anthony Hartshorn and Saghar Hosseini and Rui Hou and Hakan Inan and Marcin Kardas and Viktor Kerkez and Madian Khabsa and Isabel Kloumann and Artem Korenev and Punit Singh Koura and Marie-Anne Lachaux and Thibaut Lavril and Jenya Lee and Diana Liskovich and Yinghai Lu and Yuning Mao and Xavier Martinet and Todor Mihaylov and Pushkar Mishra and Igor Molybog and Yixin Nie and Andrew Poulton and Jeremy Reizenstein and Rashi Rungta and Kalyan Saladi and Alan Schelten and Ruan Silva and Eric Michael Smith and Ranjan Subramanian and Xiaoqing Ellen Tan and Binh Tang and Ross Taylor and Adina Williams and Jian Xiang Kuan and Puxin Xu and Zheng Yan and Iliyan Zarov and Yuchen Zhang and Angela Fan and Melanie Kambadur and Sharan Narang and Aurelien Rodriguez and Robert Stojnic and Sergey Edunov and Thomas Scialom},
      year={2023},
      eprint={2307.09288},
      archivePrefix={arXiv},
      primaryClass={cs.CL},
      url={https://arxiv.org/abs/2307.09288}, 
}

@inproceedings{cao2022machine,
  title={Machine unlearning method based on projection residual},
  author={Cao, Zihao and Wang, Jianzong and Si, Shijing and Huang, Zhangcheng and Xiao, Jing},
  booktitle={2022 IEEE 9th International Conference on Data Science and Advanced Analytics (DSAA)},
  pages={1--8},
  year={2022},
  organization={IEEE}
}

@article{hinton2015distilling,
  title={Distilling the knowledge in a neural network},
  author={Hinton, Geoffrey and Vinyals, Oriol and Dean, Jeff},
  journal={arXiv preprint arXiv:1503.02531},
  year={2015}
}

@inproceedings{guminillm,
  title={MiniLLM: Knowledge Distillation of Large Language Models},
  author={Gu, Yuxian and Dong, Li and Wei, Furu and Huang, Minlie},
  booktitle={The Twelfth International Conference on Learning Representations}
}

@inproceedings{qifine,
  title={Fine-tuning Aligned Language Models Compromises Safety, Even When Users Do Not Intend To!},
  author={Qi, Xiangyu and Zeng, Yi and Xie, Tinghao and Chen, Pin-Yu and Jia, Ruoxi and Mittal, Prateek and Henderson, Peter},
  booktitle={The Twelfth International Conference on Learning Representations}
}
